\documentclass{article}
\usepackage{spconf,amsmath,graphicx}

\usepackage{graphicx}
\usepackage{amsmath}
\usepackage{amssymb}
\usepackage{booktabs}
\usepackage{multirow}
\usepackage{enumitem}
\usepackage{subfloat}
\usepackage{subcaption}
\usepackage{bm}

\usepackage{amsmath,amsfonts}
\usepackage{algorithmic}
\usepackage{array}
\usepackage{textcomp}
\usepackage{stfloats}
\usepackage{verbatim}

\usepackage{multirow}
\usepackage{booktabs} 
\usepackage{makecell}
\usepackage{amssymb}
\usepackage{times}
\usepackage{soul}
\usepackage{url}
\usepackage[hidelinks]{hyperref}
\usepackage[utf8]{inputenc}
\usepackage{float}
\usepackage{amsmath}
\usepackage{multirow}
\usepackage{amsthm}
\usepackage{balance}
\usepackage{booktabs}
\usepackage{algorithm}
\usepackage{algorithmic}
\usepackage{pbox}

\usepackage{amsmath}
\usepackage{amssymb}
\usepackage{booktabs}
\usepackage{caption}
\usepackage{tikz}

\usepackage{colortbl}  
\usepackage{array}   

\title{Exploring vision transformer layer choosing for semantic segmentation}
\name{Fangjian Lin$^{1}$, Yizhe Ma$^{1}$, ShengWei Tian$^{1\dag}$\thanks{$\dag$ Corresponding author.}}
\address{School of Software, Xinjiang University, Urumqi, China$^1$}

\begin{document}

\maketitle

\begin{abstract}
Extensive work has demonstrated the effectiveness of Vision Transformers. The plain Vision Transformer tends to obtain multi-scale features by selecting fixed layers, or the last layer of features aiming to achieve higher performance in dense prediction tasks. However, this selection is often based on manual operation. And different samples often exhibit different features at different layers (e.g., edge, structure, texture, detail, etc.). This requires us to seek a dynamic adaptive fusion method to filter different layer features. In this paper, unlike previous encoder and decoder work, we design a neck network for adaptive fusion and feature selection, called ViTController. We validate the effectiveness of our method on different datasets and models and surpass previous state-of-the-art methods. Finally, our method can also be used as a plug-in module and inserted into different networks.
\end{abstract}


\section{Introduction}
Semantic segmentation aims at dividing each pixel of an image into a corresponding semantic class. As a fundamental research, it is widely used in many fields, such as autonomous driving, augmented reality, etc. 

The encoder and decoder architectures have been the cornerstone of semantic segmentation since FCN\cite{FCN} was proposed by Long et al. 
Recent state-of-the-arts models\cite{mask2former, segvit, StructToken} consider using a more powerful backbone\cite{ViT, Swin} and a heavier head to obtain more outstanding segmentation, but ignore the neck part that connects the backbone to the head. Some work\cite{ViTDet, ViTAdapter} has achieved more surprising results in some of the most popular methods by cleverly designing the neck connection. However, we found that recent semantic segmentation methods select layers of the backbone in a rather crude or manual way, as shown in Figure 1, \cite{ViTAdapter, StructToken, segvit} by selecting fixed layers in the backbone to act as feature map inputs at different scales. \cite{ViTDet} considers that ViT itself already has sufficient spatial information, and directly selects the features of the last layer to obtain multi-scale features.
However, in a multi-layer backbone, for different samples, there are differences in what is characterized by different layers of feature maps (e.g., edge, structure, texture, detail, etc.), and it becomes an important issue to choose which layer of features to fuse. Therefore, we want to make the model learn a more general representation, eliminate artificial bias as much as possible, and achieve better segmentation results.

In this paper, unlike previous studies (e.g., designing the encoder or designing the decoder part), we design the neck part of the segmentation model, called ViTController, to obtain a more general representation by dynamically fusing information from different layers of the feature map.
To verify the effectiveness, we evaluate our ViTController on two widely-used semantic segmentation datasets, including ADE20K \cite{ADE20K} and Cityscapes \cite{Cityscapes}, achieving 54.72$\%$, and 82.66$\%$ mIoU respectively, outperforming the state-of-the-art methods.

\begin{figure}[tp]
\centering
\includegraphics[width=1\linewidth]{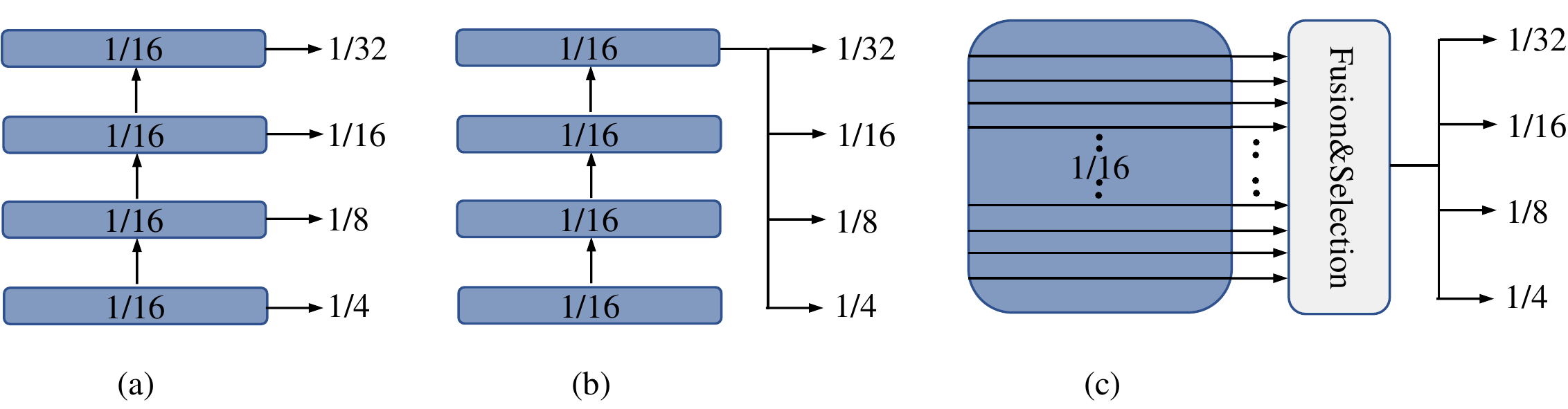} 
\caption{
Comparison with different neck mechanisms in ViT\cite{ViT}.
(a) Fixed layer feature selection.
(b) Last layer feature selection.
(c) Ours, dynamic full layer feature selection.
}
\label{diff1} 
\end{figure}

\section{related work}
In this section we focus on the layer selection of the vision transformer in semantic segmentation.
\subsection{Vision Transformer In Semantic Segmentation}
The recent Vision Transformer\cite{ViT, mae, beit} has surpassed CNN architectures in the field of vision. Although the ViT architecture performs well in semantic segmentation, it lacks the output of multi-scale feature maps (i.e., multi-scale features tend to perform better in dense prediction tasks \cite{FPN, upernet}) due to its straightforward architecture (i.e., it outputs single-scale features). \cite{beit, ViTAdapter} obtains multi-scale features by performing pooling, and de-convolution on the features of the fixed layer.  ViTDet\cite{ViTDet} directly operates on the last level of features to obtain multi-scale features.

Unlike the above manual fixed selection of features of certain layers to obtain multi-scale features, we propose a dynamic fusion approach, which adaptively fuses and controls the features of all layers, to obtain the more general representations needed for multi-scale features.

\section{Method}
\begin{figure}[!h]
\centering
\includegraphics[width=1\linewidth]{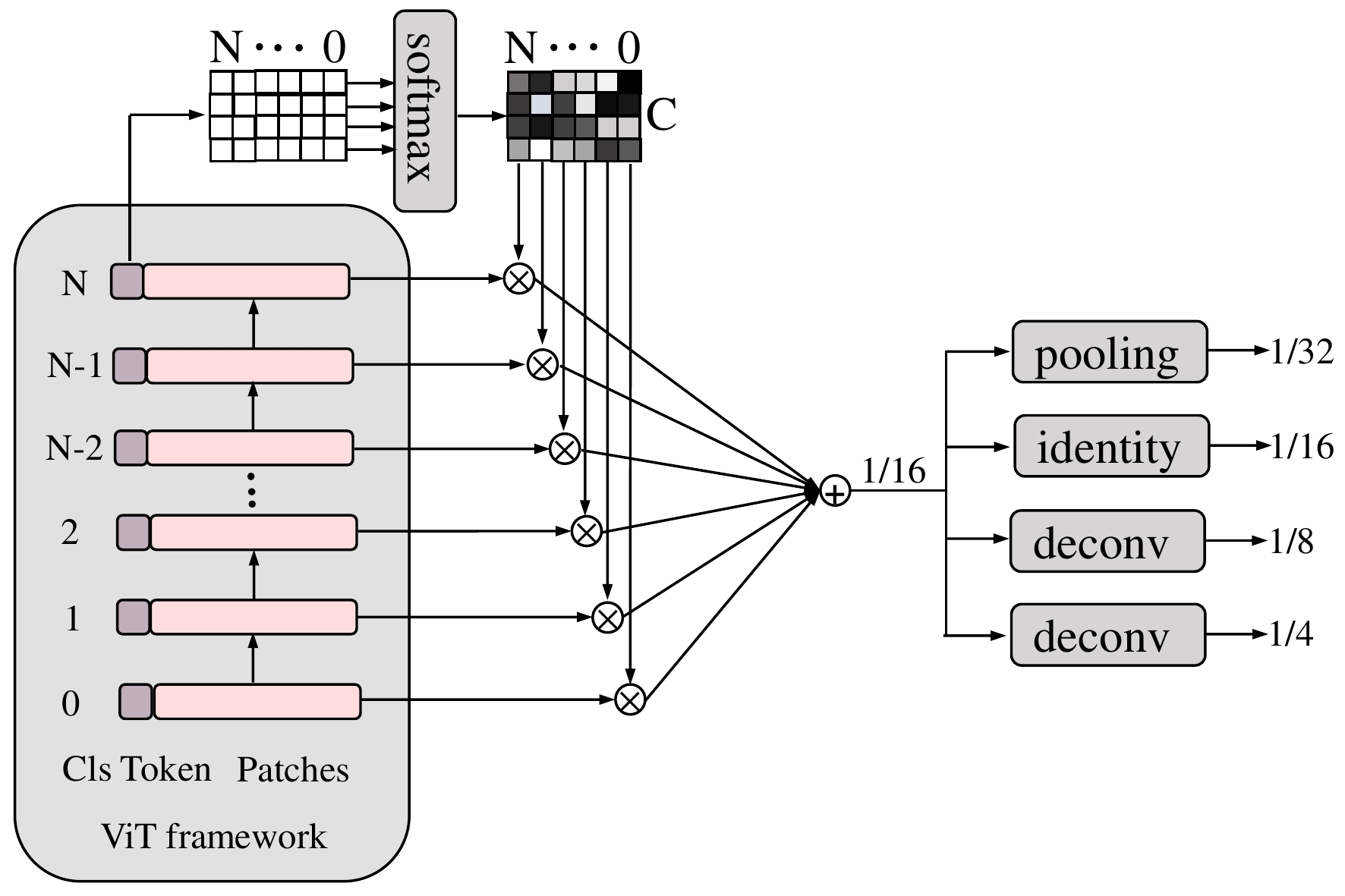} 
\caption{
The overall architecture of our ViTController. ViT framework means the plain vision transformer with class token. The right part represents the multi-scale features of the output, which is used to access the decoder head (e.g., SETR-MLA, UperNet, etc.).
}
\label{framework} 
\end{figure}

In order to get a more general representation to obtain multi-scale features, we consider modeling the all layers features of the backbone. However, the computational overhead of directly modeling the features of all layers is unaffordable. A novel idea is to leverage the class token\cite{ViT} to control the weight of all layers. Therefore, we developed a neck component with adaptive fusion and selection of all layer features.

Specifically, given an input image $X\in\mathcal{R}^{(H \times W + 1) \times 3}$, We first use ViT\cite{ViT} as Encoder to extract the features $X_i\in\mathcal{R}^{(\frac{H}{16} \times \frac{W}{16} + 1) \times C}$, where ``i'' represents the number of layers and ``1'' means class token. ``H'', ``W'' and ``C'' denote width, height and channel dimension respectively. We use cls token to obtain the feature matrix M. So we connect the class token of the N layer together. 
\begin{align}
& M = Concat(X_{cls}^{0}, X_{cls}^{2}...X_{cls}^{N}) \in \mathcal{R}^{N \times C},
\end{align}
our key idea is to get the weights of feature maps between different layers and channels to adaptive fusion and selection. Therefore, it is important to note that we need to perform softmax in the layers direction of the feature matrix M (i.e., in the ``N'' direction) to obtain the weight matrix $\hat{M}$:
\begin{align}
& \hat{M} = SoftMax(M, dim=0) \in \mathcal{R}^{N \times C}.
\end{align}
After that we need to assign the obtained weights, specifically, first expand the weights of the corresponding layer to the size of the feature map, and then perform a dot product with the feature map. This process can be described as follows:
\begin{align}
& \hat{M^0}...\hat{M^N}  = Split(\hat{M}) \in \mathcal{R}^{1 \times C},\\
& \tilde{M^0}...\tilde{M^N}  = Expand(\hat{M^0}...\hat{M^N}) \in \mathcal{R}^{\frac{H}{16} \times \frac{W}{16} \times C},\\
& \tilde{X_i}  = X_i  \tilde{M^i} \in \mathcal{R}^{\frac{H}{16} \times \frac{W}{16} \times C}.
\end{align}
Where ``$X_i$'' represents the feature after the removal of the class token, and ``i'' $\in \{0,1...N\}$. Then we connect the features for fusion:
\begin{align}
& Y  = \tilde{X_1} + \tilde{X_2} + ... + \tilde{X_N}.
\end{align}
Finally, we follow the previous approach\cite{beit, ViTAdapter} and convert the features Y to multi-scale features.

\section{Experiments}
We first introduce the datasets, implementation details, and experimental environment.
Then, we compare our method with the recent state-of-the-art on two challenging semantic segmentation benchmarks. 
Finally, we have conducted extensive ablation studies to validate the effectiveness of our method.

\subsection{Datasets}
\noindent\textbf{ADE20K}\cite{ADE20K} including 150 categories and diverse scenes with 1,038 image-level labels, which are split into 20000 and 2000 images for training and validation.

\noindent\textbf{Cityscapes}\cite{Cityscapes} contains 5K finely annotated images, split into 2975 and 500 for training and validation.

\subsection{Implementation details}
We employ ViT \cite{ViT} pretrained on ImageNet as the encoder. 
We follow the pervious training strategy\cite{Swin}: (1) random horizontal flipping, (2) random resize with a ratio between 0.5 and 2, (3) random cropping ($512 \times 512$ for ADE20K and $768\times 768$ for Cityscapes). We use the optimizer of AdamW and set the initial learning rate at 0.00002 on ADE20K and Cityscapes. The total iterations are set to 160k and 80k for ADE20K and Cityscapes, respectively. For the multi-scale validation part, we use a scale ratio of (0.5, 0.75, 1.0, 1.25, 1.5, 1.75). We use the standard mean intersection of union (mIoU) as the metric, and the widely used cross-entropy loss as the loss function.

\subsection{Reproducibility}
All our experiments use 8 NVIDIA Tesla V100 GPUs with a 32 GB memory per-card for training and inference. Experimental environment using python (version $\ge$ 3.6) and pytorch (version $\ge$ 1.7). We use MMSegmentation\cite{mmseg2020} as the training and inference framework for all experiments.

\subsection{Comparisons with the state-of-the-arts on ADE20K}
Our method uses ViT-Large\cite{ViT} as the backbone, which pretrain on ImageNet-21K, and uses SETR-MLA\cite{SETR} as the decoder head, as shown in Table \ref{sotaade}, our ViTController surpasses various most popular methods. 
\begin{table}[!h]
\setlength\tabcolsep{0.9pt}
\centering
\caption{Comparison with the state-of-the-art methods on the ADE20K dataset. ``SS'' and ``MS'' indicate single-scale inference and multi-scale inference, respectively. Flops are calculated with a resolution of 640$\times$640.}
    \label{sotaade}
  \begin{tabular}{l|c|c|c|c|c}
    \hline
    \hline
    {Method} & {Backbone} & {GFLOPs} & {Params} & {SS} & {MS}\\
    \hline
    DPT\cite{DPT}              & ViT-L/16      &328 &338M &49.16 & 49.52\\
    UperNet\cite{upernet}      & ViT-L/16      &710 &354M &48.64 & 50.00\\
    SETR \cite{SETR}           & ViT-L/16      &332 &310M &50.45 & 52.06\\
    MCIBI \cite{MCIBI}         & ViT-L/16      &-   &-    &-     & 50.80\\
    Segmenter \cite{Segmenter} & ViT-L/16      &380 &342M &51.80 & 53.60\\
    StructToken\cite{StructToken} & ViT-L/16      &398 &350M &52.84 & 54.18\\
    ViTController (ours)            & ViT-L/16      &332&310M&\textbf{53.35} & \textbf{54.72}\\
    \hline
    \hline
    \end{tabular}
\end{table}
It can be seen that our ViTController is +2.9$\%$ mIoU (50.45$\%$ vs. 53.35$\%$) higher than SETR \cite{SETR}. When multi-scale testing is adopted, our ViTController is +2.66$\%$ mIoU (52.06$\%$ vs. 54.72$\%$) higher than SETR. In the previous approach, StructToken\cite{StructToken} achieved the best performance, our method is +0.51$\%$ mIoU (52.84$\%$ vs. 53.35$\%$) higher than it, and +0.54$\%$ mIoU (54.18$\%$ vs. 54.72$\%$) on multi-scale testing. In addition we have conducted extensive ablation studies on multiple decoder heads\cite{SETR, upernet, semanticfpn} with multi-scale inputs, please refer to Table \ref{ablation2} for details.

\subsection{Comparisons with the state-of-the-arts on Cityscapes}
Cityscapes is a large resolution segmentation dataset that contains many objects of varying sizes. In this area, ViT is mediocre, but our approach has improved.
\begin{table}[!h]
\setlength\tabcolsep{0.9pt}
\centering
\caption{Comparison with the state-of-the-art methods on the Cityscapes validation set. Flops are calculated with a resolution of 768$\times$768.}
  \begin{tabular}{l|c|c|c|c|c}
    \hline
    \hline
    {Method} & {Backbone} & {GFLOPs} & {Params} & {SS} & {MS}\\
    \hline
    Segmenter \cite{Segmenter}   & ViT-L/16      &553 &340M &79.10  & 81.30\\
    StructToken\cite{StructToken}    & ViT-L/16      &600 &364M &80.05 & 82.07\\
    SETR \cite{SETR} & ViT-L/16      &589 &319M &-  & 82.15\\
    ViTController (ours)            & ViT-L/16      &590&320M&\textbf{81.05} & \textbf{82.66}\\
    \hline
    \hline
    \end{tabular}
    \label{sotacitys}
\end{table}
As shown in Table \ref{sotacitys}, our method is +0.51$\%$ mIoU (82.15$\%$ vs. 82.66$\%$) higher than SETR when multi-scale testing is adopted.

\subsection{Insert our method into the different decoder heads}
To clearly demonstrate the effectiveness of our method, we insert our method into several decoder methods that require multi-scale input. For the efficiency of the experiments, we used ViT-Small as the encoder for all ablation studies. 
\begin{table}[!h]
\setlength\tabcolsep{2pt}
\centering
\caption{The efficiency of our approach on different methods.}
  \begin{tabular}{l|c|l|l}
    \hline
    \hline
    {Method} & {Backbone} & {SS} & {MS}\\
    \hline
    Semantic FPN\cite{semanticfpn}        & ViT-S/16      &44.80              & 45.91\\
    Semantic FPN + ours & ViT-S/16      &47.28        & 49.14 \\
    \hline
    UperNet\cite{upernet}             & ViT-S/16      &45.53              & 46.14\\
    UperNet + ours      & ViT-S/16      &47.88        & 49.17\\
    \hline
    SETR-MLA\cite{SETR}                & ViT-S/16      &44.85              & 46.30\\
    SETR-MLA + ours         & ViT-S/16      &47.60        & 49.51\\
    
    \hline
    \hline
    \end{tabular}
    \label{ablation2}
\end{table}
As shown in Table \ref{ablation2}, when plugging our method into the plain Semantic FPN\cite{semanticfpn}, it can be seen that single-scale and multi-scale are +2.48$\%$ mIoU (44.80$\%$ vs. 47.28$\%$) and +3.23$\%$ mIoU (45.91$\%$ vs. 49.14$\%$), respectively. UperNet\cite{upernet} is used as a baseline method for performance evaluation by many Transformer methods. Our method is + 2.35$\%$ mIoU (45.53$\%$ vs. 47.88$\%$) higher than it. For multi-scale testing, our method is + 3.03$\%$ mIoU (46.14$\%$ vs. 49.17$\%$) higher than UperNet. SETR-MLA\cite{SETR} is a lightweight decoder head, which achieves trade-offs in terms of efficiency and performance. Based on the SETR-MLA head, our method improves 3.21$\%$ mIoU and achieves 49.51$\%$ mIoU when ViT-Small is used as the backbone.

\subsection{Comparison with different control manner}
Direct global modeling of full-layer features would result in a huge computational overhead. A simple idea is to obtain general representations by using a global average pooling operation, and by modeling the global feature vector to obtain the weights at different levels. As shown in Table \ref{ablation1}, Average Pooling operation is +1.85$\%$ mIoU (46.30$\%$ vs. 48.15$\%$) higher than baseline.

\begin{table}[!h]
\setlength\tabcolsep{4pt}
\centering
\caption{Ablation study on different ways of composing the weight matrix. Our baseline method consists of ViT-Small and SETR-MLA head.}
  \begin{tabular}{l|c|c|c}
    \hline
    \hline
    {Method} & {Backbone} & {SS} & {MS}\\
    \hline
    baseline              & ViT-S/16      &44.85          & 46.30\\
    Average Pooling       & ViT-S/16      &46.79          & 48.15\\
    Class Token           & ViT-S/16      &47.60          & 49.51\\
    \hline
    \hline
    \end{tabular}
    \label{ablation1}
\end{table}
In addition, the class token was proposed as an image classifier by Dosovitskiy et al. Considering the specificity of the class token in ViT (i.e., pretrained on ImageNet-1k, global representation of features and no inductive bias), we consider using the class token as a controller for feature fusion and selection. It can be seen that the Class Token is +1.36$\%$ mIoU (48.15$\%$ vs. 49.51$\%$) higher than the Average Pooling operation. Therefore we choose the class token as the controller for the feature fusion and selection for all experiments.

\subsection{Visual Analysis}
\begin{figure*}[tp]
\centering

\includegraphics[width=1\linewidth]{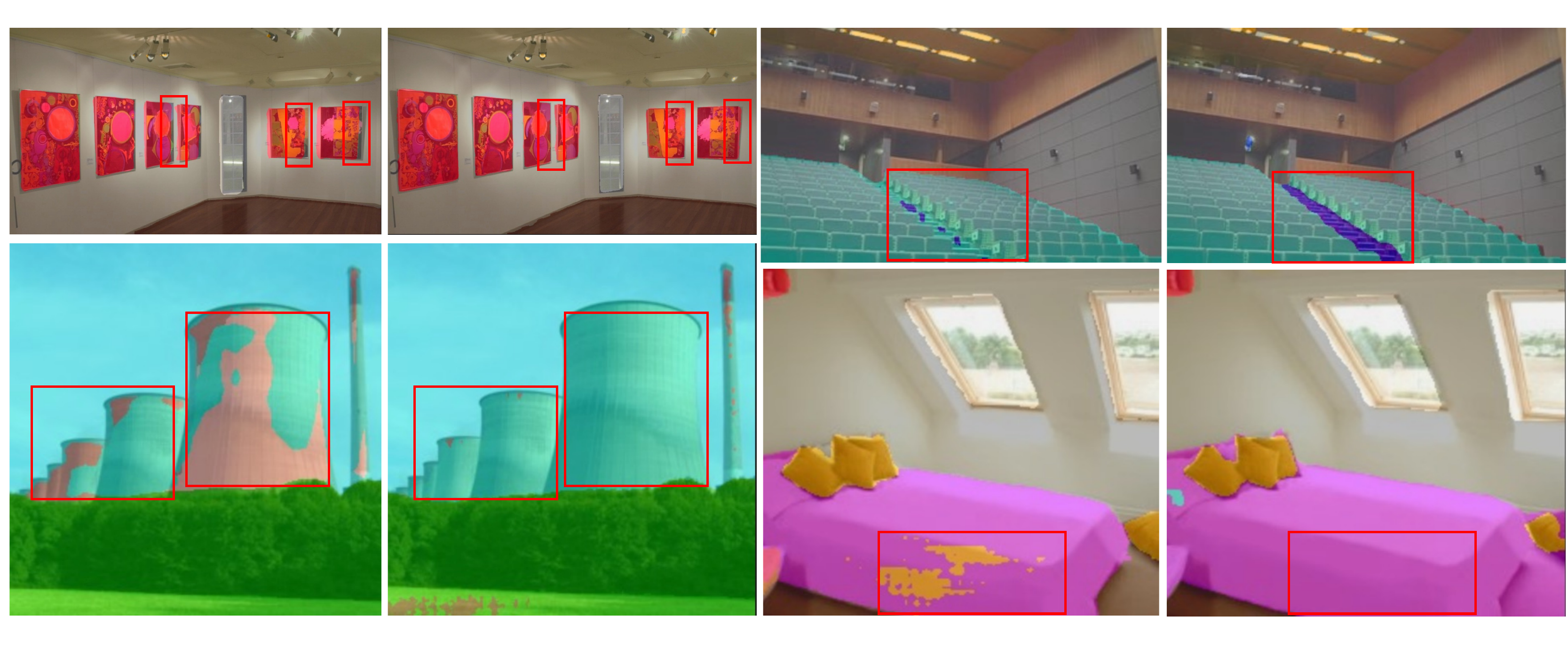} 
\caption{Qualitative results on ADE20k: Ours (right column) vs. SETR-MLA (baseline method, left column) in each pair. Best viewed in
color and zoom in.}
\vspace{0.5cm}
\label{vis}
\end{figure*}

Visualization is shown in Figure \ref{vis}. We compared ViTController with the baseline model, it can be found that we not only handle the global features better but also perform well in capturing local details.

\section{Limitation}
When dealing with scenes like Cityscapes where objects are of different sizes, Vision Transformer\cite{ViT} is still not as effective as CNN networks. Such scenarios often require models with excellent local information extraction and multi-scale context modeling capabilities.
But the plain Vision Transformer lacks scale variation, so we want to enhance the multi-scale context fusion in the neck part to cope with the scenes with changing size objects and achieve better segmentation. And our method considers different layers of feature selection and lacks multi-scale context modeling, so we hope to make progress in future work.
\section{Conclusion}
In this paper, we eliminate the inductive bias caused by the manual selection of fixed-layer features by introducing a dynamic adaptive feature fusion and selection controller. With a simple and efficient Neck network, it makes the output representation more general, maximizing the expectation and weight assignment for different samples.

We validate the effectiveness of our method on two widely used semantic segmentation datasets and surpass the previous sota methods. Finally, we hope to shift the researcher's attention to the design of the neck network, making the neck network can connect the backbone and the decoder head in a silky smooth way.

\section{Acknowledgments}
The study is partially supported by the National Natural Science Foundation of China under Grant (U2003208) and Key R \& D Project of Xinjiang Uygur Autonomous Region(2021B01002).

\clearpage
\bibliographystyle{IEEEbib}
\bibliography{strings,refs}

\end{document}